# Why We Do Not Evolve Software? Analysis of Evolutionary Algorithms

**Roman V. Yampolskiy**
Computer Engineering and Computer Science
University of Louisville
roman.yampolskiy@louisville.edu

**Abstract**
In this paper, we review the state-of-the-art results in evolutionary computation and observe that we don't evolve non-trivial software from scratch and with no human intervention. A number of possible explanations are considered, but we conclude that computational complexity of the problem prevents it from being solved as currently attempted. A detailed analysis of necessary and available computational resources is provided to support our findings.

*"We point out a curious philosophical implication of the algorithmic perspective: if the origin of life is identified with the transition from trivial to non-trivial information processing – e.g. from something akin to a Turing machine capable of a single (or limited set of) computation(s) to a universal Turing machine capable of constructing any computable object (within a universality class) – then a precise point of transition from non-life to life may actually be undecidable in the logical sense. This would likely have very important philosophical implications, particularly in our interpretation of life as a predictable outcome of physical law."* [1].

## 1. Introduction

On April 1, 2016 Dr. Yampolskiy posted the following to his social media accounts: "Google just announced major layoffs of programmers. Future software development and updates will be done mostly via recursive self-improvement by evolving deep neural networks". The joke got a number of "likes" but also, interestingly, a few requests from journalists for interviews on this "developing story". To non-experts the joke was not obvious, but why? Why don't we evolve software? A quick search produced no definitive answers, and so this paper was born.

In 1859 Charles Darwin [2] and many scholars before him [3, 4] have proposed theories to explain the origins of complex life forms via natural selection and modification. Scientific theories are algorithms which given as input starting conditions make statistically accurate predictions of the future state of the system. For example, computer simulations of continental drift give us positions of continents at some time t. Yampolskiy emphases importance of such simulations, "A scientific theory cannot be considered fully accepted until it can be expressed as an algorithm and simulated on a computer. It should produce observations consistent with measurements obtained in the real world, perhaps adjusting for the relativity of time scale between simulation and the real world. In other words, an unsimulatable hypothesis should be considered significantly weaker than a simulatable one. It is possible that the theory cannot be simulated due to limits in our current computational capacity, hardware design, or capability of programmers and that it will become simulatable in the future, but until such time, it should have a tentative status." [5].

Simulations of Darwinian algorithm on a computer are known as Evolutionary Algorithms (EA) and have been around since the early days of computer science [6, 7], with popular sub-fields such as Genetic Algorithms (GA), Genetic Programming (GP), Evolutionary Strategy (ES) and Artificial Life (AL). Currently, the state of performance in all of the above-mentioned areas is orders of magnitude less complex than what we observe in the natural world, but why?

A number of seminal papers have been published attempting to formalize Darwin's biological theory from the point of view of computational sciences. Such works essentially see biological evolution as a computational process running on a carbon-based substrate, but which has computationally equivalent algorithms in other substrates. Valiant in his work on evolvability [8] treats Darwinian evolution as a learning process over mathematical functions and attempts to explain quantitatively which artifacts can be evolved with given resources, and which can not. Likewise, Chaitin in his work on metabiology [9, 10] attempts to develop an abstract fundamental mathematical theory of evolution. Wolfram in his, "a New Kind of Science" [11], attempts to show how rules of computational universe of simple programs can be used to explain some of the biological complexity we observe. Livnat and Papadimitriou analyze sex as an algorithm, in their work on the theory of evolution viewed through the lens of computation [12].

It is interesting to do a thought experiments and try to imagine what testable predictions Charles Darwin would have made, had he made his discovery today, with full knowledge of modern bioinformatics and of computer science. His predictions may have included the following: 1) Simulations of evolution will produce statistically similar results at least with respect to complexity of artifacts produced. 2) If running evolutionary algorithms for as long as possible continued to produce non-trivial outputs, scientists would run them forever. Likewise, he would be able to make some predictions, which would be able to falsify his theory, such as: 1) Representative simulations of evolution will not produce similar results to those observed in nature. 2) Researchers will not be able to evolve software or other complex or novel artifacts. 3) There will not be any projects running evolutionary algorithms long-term because their outputs would quickly stop improving and stabilize. With respect to the public and general cultural knowledge, it would be reasonable to predict that educated people would know the longest-running evolutionary algorithm, and the most complex evolved algorithm. Similarly, even schoolchildren would know the most complex digital organism ever evolved.

In the rest of the paper we evaluate the state-of-the-art in relevant published research to see how above mentioned predictions and counterfactuals hold up and what we can say about the foundational question of this paper. We analyze potential explanations for the current observations of progress in the domain of EAs and look at computational resources, required and available, as the main source of limitations and future opportunities for evolving software.

## 2. Evolutionary Computation

Inspired by the Darwin's theory [2] of biological evolution, evolutionary computation attempts to automate the process of optimization and problem solving by simulating differential survival and reproduction of individual solutions. From the early 1950s multiple well documents attempts to make Darwin's algorithm work on a computer have been published under such names as Evolutionary Programming [13], Evolutionary Strategies [14], Genetic Algorithms [15], Genetic Programming [16], Genetic Improvement [17], Gene Expression Programming [18], Differential Evolution [19], Neuroevolution [20] and Artificial Embryogeny [21]. While numerous variants different in their problem representation and metaheuristics exist [22-25], all can be reduced to just two main approaches – Genetic Algorithm (GA) and Genetic Programming (GA).

GAs are used to evolve optimized solutions to a particular instance of a problem such as Shortest Total Path [26], Maximum Clique [27], Battleship [28], Sudoku [29], Mastermind [24], Light Up [30], Graph Coloring [31], integer factorization [32, 33] or efficient halftone pattern for printing [34] and so are not the primary focus of this paper. GPs purpose, from their inception, was to automate programming by evolving an algorithm or a program for solving a particular class of problems, for example an efficient [35] search algorithm. Software design is the type of application most frequently associated with GPs [36], but work in this domain is also sometimes referred to as "real programing", "object-oriented GP", "traditional programming", "Turing Equivalent (TE) programming" or "Turing-complete GP" [37, 38].

The sub-field of computation inspired by evolution in general and Genetic Programing paradigm, established by John Koza in 1990s , in particular are thriving and growing exponentially as evidenced both by the number of practitioners and of scientific publications. Petke et al. observe "…enormous expansion of number of publications with the Genetic Programming Bibliography passing 10,000 entries … By 2016 there were nineteen GP books including several intended for students …" [17]. Such tremendous growth has been fueled, since early days, by belief in capabilities of evolutionary algorithms, and our ability to overcome obstacles of limited compute or data as illustrated by the following quotes:

- "We will (before long) be able to run genetic algorithms on computers that are sufficiently fast to recreate on a human timescale the same amount of cumulative optimization power that the relevant processes of natural selection instantiated throughout our evolutionary past … " [39]
- "As computational devices improve in speed, larger problem spaces can be searched." [40].
- "We believe that in about fifty years' time it will be possible to program computers by means of evolution. Not merely possible but indeed prevalent." [41]. "The relentless iteration of Moore's law promises increased availability of computational resources in future years. If available computer capacity continues to double approximately every 18 months over the next decade or so, a computation requiring 80 h will require only about 1% as much computer time (i.e., about 48 min) a decade from now. That same computation will require only about 0.01% as much computer time (i.e., about 48 seconds) in two decades. Thus, looking forward, we believe that genetic programming can be expected to be increasingly used to automatically generate ever-more complex human-competitive results." [42].
- "The production of human-competitive results as well as the increased intricacy of the results are broadly correlated to increased availability of computing power tracked by Moore's law. The production of human-competitive results using genetic programming has been greatly facilitated by the fact that genetic algorithms and other methods of evolutionary computation can be readily and efficiently parallelized. … Additionally, the production of human-competitive results using genetic programming has facilitated to an even greater degree by the increased availability of computing power, over a period of time, as tracked by Moore's law. Indeed, over the past two decades, the number and level of intricacy of the human-competitive results has progressively grown. … there is, nonetheless, data indicating that the production of human-competitive results using genetic programming is broadly correlated with the increased availability of computer power, from year to year, as tracked by Moore's Law." [42].
- "… powerful test data generation techniques, an abundance of source code publicly available, and importance of nonfunctional properties have combined to create a technical and scientific environment ripe for the exploitation of genetic improvement." [40].

## 3. State-of-the-Art with Respect to Predictions

In order to establish the state-of-the-art in evolutionary computation we examined a number of survey papers [42, 43] and seminal results [44-49] looking at produced human-competitive results, as they are meant to represent the greatest accomplishments of the field. While, on the surface the results may seem impressive, deeper analysis shows complete absence of success in evolving non-trivial software from scratch and without human assistance. It is of course necessary to be precise about what it is we are trying to measure or detect, as to avoid disagreements resulting from ambiguity in terms being used.

It may be difficult to formally specify what makes a piece of software non-trivial, but intuitively-attractive measure of length of the program expressed as the number of lines of code is not a sufficient indicator of complexity, as it could have an extremely low Kolmogorov complexity [50]. Inspired by the Turing Test [51, 52], which is based on inability to distinguish output from a person and a computer, we propose defining non-trivial software as such which would take an average experienced human programmer at least a full hour of effort to produce if given the same problem to solve. If the solution source code could be produced with significantly less effort (ex. 1 minute), it may not be sufficiently complex and the problem may be deemed trivial for our purposes. Our approach to specifying triviality would exclude "Hello World" and most other toy programs/problems from consideration, which is exactly what we wanted to achieve as the main benefit from being able to evolve software would come from ability to replace full time programmers.

With regards to the other two conditions, they are much easier to specify. From "scratch" means that we are not starting with an existing version of a program (but are happy to rely on existing APIs, subject to the non-triviality of all newly produced code). Without human assistance can be interpreted to mean that the programmer is working alone, or a team of programmers is working an equivalent amount of time, for example two programmers would each need at least 40 minutes to solve the problem, which implies a small communication overhead.

Reading early claims about capabilities of EA feels just like reading early predictions from AI literature [53]. Some early success is projected into the future by assuming that the same rate of progress with continue and it is claimed that complete success is just years away. However, just like with early AI, the claims are inflated, unsupported, overly optimistic, phrased in deceptive and metaphoric language, and the solutions do not scale to the real world problems. Perhaps an

EA "winter" is long overdue. Here is how Koza, presents the state of the field in 1994: "… in this article, we will present a number of examples from various fields supporting the surprising and counter-intuitive notion that computers can indeed by programmed by means of natural selection. We will show, via examples, that the recently developed genetic programming paradigm provides a way to search the space of all possible programs to find a function which solves, or approximately solves, a problem." [16].

Sixteen years later he reports results of what he calls an 'extraordinary long experiment': "An additional order-of-magnitude increase was achieved by making extraordinarily long runs on the largest machine (the 1,000-node machine). … The length of the run that produced the two patentable inventions was 28.8 days—almost an order-of-magnitude increase (9.3 times) over the overall 3.4-day average for typical runs of genetic programming that our group had been making at the time." [42]. One quickly realizes that most improvements in the field simply come from using more compute to search progressively larger parts of the solutions space, a result similar to the one expected for random search algorithm.

Here is an example of overhyped and ambiguous reporting of results, this time from recent work on EA. Researchers Becker and Gottschlich [40] go from naming their paper - "AI Programmer: Autonomously Creating Software Programs Using Genetic Algorithms" to abstract "AI Programmer, that can automatically generate full software programs requiring only minimal human guidance." To claiming that "Using AI Programmer, we were able to generate numerous complete software programs." Finally in experimental results they state what they managed to produce "A generated program that outputs 'hello' " or performs addition operation. [40]. But even that is a bit of a hype, "Rather than starting with "Hello World", we first had AI Programmer create a more simplistic program that simply output "hi." It was successfully after 5,700 generations…" [40]. Even this trivial one-liner was not a clean success. "The generated program fulfilled its requirement to output the target text, but interestingly included subsequent random characters, which contained parsing errors, including nonmatching brackets." [40]. An identical program but the one printing "I love all humans" took 6,057,200 generations. [40].

Perhaps it is unfair to pick on this particular paper, which is only available as an unreviewed pre-print, but we selected it because it is highly representative of the type of work frequently published in GP, and its extremeness makes problems clear to identify. If its title was "Brute Forcing Strings" it would be a reasonable work on that subject, but like so many others authors claim to "Autonomously Creating Software Programs" using evolutionary computation, a claim which is never substantiated in any published literature on this subject. We are not alone in our skepticism; many others have arrived at exactly the same conclusions:

- In practice however, GPs are used in the same way as GAs, for optimization of solutions to particular problems or for function optimization [37, 38, 54-57] or for software improvement [58].
- "We examine what has been achieved in the literature, and find a worrying trend that largely small toy-problems been attempted which require only a few line of code to solve by hand." [38].
- "A literature review has revealed that only a small fraction of the papers on GP deal with evolving TE computer programs, with the ability to iterate and utilize memory, while the majority of papers deal with evolving less expressive logical or arithmetic functions." [38]. "We conclude that GP in its current form is fundamentally awed, when applied to the space of TE programs." [38]. "Computer code is not as robust as genetic code, and therefore poorly suited to the process of evolution, resulting in a insurmountable landscape which cannot be navigated effectively with current syntax based genetic operators. Crossover, has problems being adopted in a computational setting, primarily due to a lack of context of exchanged code. A review of the literature reveals that evolved programs contain at most two nested loops, indicating that a glass ceiling to what can currently be accomplished." [38].

- "A full understanding of open-ended evolutionary dynamics remains elusive" [59].
- "There are many problems that traditional Genetic Programming (GP) cannot solve, due to the theoretical limitations of its paradigm. A Turing machine (TM) is a theoretical abstraction that express the extent of the computational power of algorithms. Any system that is Turing complete is sufficiently powerful to recognize all possible algorithms. GP is not Turing complete." [57].

Even a survey of GP community itself produced the following feedback regarding current problems being worked on:

• "Far too many papers include results only on simple toy problems which are often worse than meaningless: they can be misleading";
• "(we should exclude) irrelevant problems that are at least 20 years old";
• "get rid of some outdated, too easy benchmarks";
• "the standard 'easy' Koza set should not be included"
• "[it is] time to move on". [37].

With regards to Darwin's hypothetical predictions raised in the introduction we can state the following:

**Prediction**. Simulations of evolution will produce statistically similar results at least with respect to complexity of artifacts produced.
**Status.** False as of 2018.

**Prediction.** If running evolutionary algorithms for as long as possible continued to produce non-trivial outputs, scientists would run them forever.
**Status.** False as of 2018.

**Prediction.** Representative simulations of evolution will not produce similar results to those observed in nature.
**Status.** True as of 2018.

**Prediction.** Researchers will not be able to evolve software or other complex or novel artifacts.
**Status.** True as of 2018.

**Prediction.** There will not be any projects running evolutionary algorithms long-term because their outputs would quickly stop improving and stabilize.
**Status.** True as of 2018.

**Prediction.** With respect to the public and general cultural knowledge, it would be reasonable to predict that educated people would know the longest-running evolutionary algorithm, and the most complex evolved algorithm.
**Status.** False and False as of 2018.

**Prediction.** Similarly, even schoolchildren would know the most complex digital organism ever evolved.
**Status.** False as of 2018.

Looking at outcomes from the made predictions we observe that all predictions are false as of 2018 and all counterfactuals are true as of the same year as long as we look only at non-trivial products of evolutionary computations. We are not evolving complex artifacts, we are not running evolutionary algorithms for as long as possible, we are not evolving software, and the public is unaware of most complex products of evolutionary computation. On close examination all "human-competitive" results turn out to be just optimizations, never fully autonomous programming leading to novel software being engineered.

## 4. Possible Explanations

A number of possible explanations for "Why we don't evolve software?" could be considered. We tried to be as comprehensive as possible in this section, but it is possible that we have not considered some plausible explanation.

- **Incompetent programmers**
  It is theoretically possible, but is highly unlikely, that out of thousands of scientists working on evolutionary computation all failed to correctly implement the Darwinian algorithm.
- **Non-representative algorithms**
  Some [57] have suggested that evolutionary algorithms do not accurately capture the theory of evolution, but of course, that would imply that the theory itself is not specified in sufficient detail to make falsifiable predictions. If on the other hand, such more detailed specifications are available to our critics, it is up to them to implement them as computer simulations for testing purposes, but no successful examples of such work is known and the known ones have not been successful in evolving software.
- **Insufficient complexity of the environment (not enough data, poor fitness functions)**
  It is possible that the simulated environment is not complex enough to generate high complexity outputs in evolutionary simulations. This does not seem correct as Internet presents a highly complex landscape in which many self-modifying computer viruses roam [60]. Likewise, virtual world such as Second Life and many others present close approximations to the real world and a certainly more complex than early Earth was. Requiring more realistic environmental conditions, may results in an increase in necessary computational resources, a problem addressed in the next bullet. "A skeptic might insist that an abstract environment would be inadequate for the evolution …, believing instead that the virtual environment would need to closely resemble the actual biological environment in which our ancestors evolved. Creating a physically realistic virtual world would require a far greater investment of computational resources than the simulation of a simple toy world or abstract problem domain (whereas evolution had access to a physically realistic real world "for free"). In the limiting case, if complete microphysical accuracy were insisted upon, the computational requirements would balloon to utterly infeasible proportions." [39].

- **Insufficient resources (Compute, Memory)**
  From the history of computer science, we know of many situations (speech recognition, NN training), where we had a correct algorithm, but insufficient computational resources to run it to success. It is possible that we simply don't have hardware powerful enough to emulate evolution. We will address this possibility in Section 5.
- **Software design is not amenable to evolutionary methods**
  Space of software designs may be discreet with no continues path via incremental fitness to the desired solutions. This is possible, but this implies that original goals of GP are unattainable and misguided. In addition, since a clear mapping exists between solutions to problems and animals as solutions to environmental problems this would also imply that current explanation for the origin of the species is incorrect [61].
- **Darwinian algorithm is incomplete or wrong**
  Lastly, we have to consider the possibility that the inspiration behind evolutionary computation, the Darwinian algorithm itself is wrong or at least partially incomplete. If that was true, computer simulations of such algorithm would fail to produce results comparable to observations we see in nature and a search for an alternative algorithm would need to take place. This would be an extremely extraordinary claim and would require that we first discard all the other possible explanations from this list first.

Perhaps, we can learn from similar historical problems. Earliest work on artificial neurons was done in 1943 by McCulloch and Pitts [62] and while research on Artificial Neural Networks (ANN) continued [63], until 2010 it would have been very logical to ask: "Why don't artificial neural networks perform as well as natural ones?" Today, deep neural networks frequently outperform their human counterparts [64, 65], but it may still be helpful to answer this question about NN, to see how it was resolved. Stuhlmüller succinctly summarizes answer given by Ghahramani: "Why does deep learning work now, but not 20 years ago, even though many of the core ideas were there? In one sentence: We have more data, more compute, better software engineering, and a few algorithmic innovations …" [66]. Consequently, the next section looks at this very-likely explanation in detail.

## 5. Computational Complexity of Biological Evolution and Available Compute

In the biological world, evolution is a very time consuming process with estimates for the appearance of early life pointing to some 4 billion years ago and each new generation taking minutes for simple life forms like bacteria and about 20 years for more complex species, like Homo Sapiens. Given the timescales involved, it is impossible to replicate full-scale evolution in experimental settings, but it may be possible to do so in computer simulations, by generating new offspring in matter of milliseconds and by greatly expediting necessary fitness evaluation time, potentially reducing a multi-billion year natural process to just a few years of simulation on a powerful supercomputer. Others have thought about the same, "*What algorithm could create all this in just $10^{12}$ steps?* The number $10^{12}$—one trillion—comes up because this is believed to be the number of generations since the dawn of life $3.5 \cdot 10^9$ years ago (notice that most of our ancestors could not have lived for much more than a day). And it is not a huge number: cellphone processors do many more steps in an hour." [12].

Hamiltonian complexity [67] studies how hard is it to simulate a physical system, where "hard" means that the computational resources required to approximate behavior of the system grow too quickly with the size of the system being simulated, so that no computer can carry out the task in reasonable time [67]. Specifically, in the context of evolutionary algorithms, research effort to establish bounds and improve efficiency is known as Evolutionary Algorithm Theory (EAT) [68]. In this section, we will attempt to estimate the computational power of evolution in biosphere, analyze computational complexity of bio-inspired evolutionary algorithms and finally compare our findings to the available and anticipated computational resources; all in the hopes of understanding, if it is possible to replicate evolution on a computer, in practice.

Similar attempts have been made by others, for example Shulman and Bostrom wanted to figure out computational requirements necessary to evolve Artificial Intelligence: "The argument from evolutionary algorithms then needs one additional premise to deliver the conclusion that engineers will soon be able to create machine intelligence, namely that we will soon have computing power sufficient to recapitulate the relevant evolutionary processes that produced human intelligence. Whether this is plausible depends both on what advances one might expect in computing technology over the next decades and on how much computing power would be required to run genetic algorithms with the same optimization power as the evolutionary process of natural selection that lies in our past. One might for example try to estimate how many doublings in computational performance, along the lines of Moore's law, one would need in order to duplicate the relevant evolutionary processes on computers." [39].

By looking at total number of generations, population sizes, DNA storage [69-72] and computation and involved neural information processing it is possible to arrive at broad estimates of computational power behind biological evolution. "In this way, the biosphere can be visualised as a large, parallel supercomputer, with the information storage represented by the total amount of DNA and the processing power symbolised by transcription rates. In analogy with the Internet, all organisms on Earth are individual containers of information connected through interactions and biogeochemical cycles in a large, global, bottom-up network." [73]. "We have various methods available to begin to estimate the power of evolutionary search on Earth: estimating the number of generations and population sizes available to human evolution, creating mathematical models of evolutionary "speed limits" under various conditions, and using genomics to measure past rates of evolutionary change. " [39].

- With respect to the estimates of the storage capabilities of the biosphere we have: "The total amount of DNA contained in all of the cells on Earth is estimated to be about $5.3 \times 10^{37}$ base pairs [73], equivalent to $1.325 \times 10^{37}$ bytes of information." [74].
- "Modern whole-organism genome analysis, in combination with biomass estimates, allows us to estimate a lower bound on the total information content in the biosphere: $5.3 \times 10^{31}$ ($\pm 3.6 \times 10^{31}$) megabases (Mb) of DNA. Given conservative estimates regarding DNA transcription rates, this information content suggests biosphere processing speeds exceeding yottaNOPS values ($10^{24}$ Nucleotide Operations Per Second)." [73].
- "Finding the amount of DNA in the biosphere enables an estimate of the computational speed of the biosphere, in terms of the number of bases transcribed per second, or Nucleotide Operations Per Second (NOPS), analogous to the Floating-point Operations Per

Second (FLOPS) metric used in electronic computing. A typical speed of DNA transcription is 18–42 bases per second for RNA polymerase II to travel along chromatin templates … and elsewhere suggested as 100 bases per second …. Precisely how much of the DNA on Earth is being transcribed at any one time is unknown. The percentage of any given genome being transcribed at any given time depends on the reproductive and physiological state of organisms, and at the current time we cannot reliably estimate this for all life on Earth. If all the DNA in the biosphere was being transcribed at these reported rates, taking an estimated transcription rate of 30 bases per second, then the potential computational power of the biosphere would be approximately $10^{15}$ yottaNOPS (yotta = $10^{24}$), about $10^{22}$ times more processing power than the Tianhe-2 supercomputer …, which has a processing power on the order of $10^5$ teraFLOPS (tera = $10^{12}$). It is estimated that at 37°C, about 25% of Open Reading Frames in Escherichia coli are being transcribed …, but this is in a metabolically active population." [73].

- To estimate neural information processing of nature we need to look at the processing power of all neurons in the biosphere: "There are some 4-6*$10^{30}$ prokaryotes in the world today, but only $10^{19}$ insects, and fewer than $10^{10}$ human (pre-agricultural populations were orders of magnitude smaller). However, evolutionary algorithms require not only variations to select among but a fitness function to evaluate variants, typically the most computationally expensive component. A fitness function for the evolution of artificial intelligence plausibly requires simulation of "brain development," learning, and cognition to evaluate fitness. We might thus do better not to look at the raw number of organisms with complex nervous systems, but instead to attend to the number of neurons in biological organisms that we might simulate to mimic evolution's fitness function. We can make a crude estimate of that latter quantity by considering insects, which dominate terrestrial biomass, with ants alone estimated to contribute some 15-20% of terrestrial animal biomass. Insect brain size varies substantially, with large and social insects enjoying larger brains; e.g., a honeybee brain has just under $10^6$ neurons, while a fruit fly brain has $10^5$ neurons, and ants lie in between with 250,000 neurons. The majority of smaller insects may have brains of only a few thousand neurons. Erring on the side of conservatively high, if we assigned all $10^{19}$ insects fruit-fly numbers of neurons the total would be $10^{24}$ insect neurons in the world. This could be augmented with an additional order of magnitude, to reflect aquatic copepods, birds, reptiles, mammals, etc., to reach $10^{25}$. (By contrast, in pre-agricultural times there were fewer than $10^7$ humans, with under $10^{11}$ neurons each, fewer than $10^{18}$ total, although humans have a high number of synapses per neuron.) The computational cost of simulating one neuron depends on the level of detail that one wants to include in the simulation. Extremely simple neuron models use about 1,000 floating-point operations per second (FLOPS) to simulate one neuron (for one second of simulated time); an electrophysiologically realistic Hodgkin-Huxley model uses 1,200,000 FLOPS; a more detailed multicompartmental model would add another 3-4 orders of magnitude, while higher-level models that abstract systems of neurons could subtract 2-3 orders of magnitude from the simple models. If we were to simulate $10^{25}$ neurons over a billion years of evolution (longer than the existence of nervous systems as we know them) in a year's run time these figures would give us a range of $10^{31}$-$10^{44}$ FLOPS." [39].

As Darwinian algorithm is inherently probabilistic, it is likely that many runs of the algorithm are required to have just one of them succeed, just like in the case of biological evolution [75]. The number of such simultaneous runs can be estimated from the total size of the search space divided by the average individual computational resources of each run. In the special case of biological evolution evolving intelligent beings, "The observation selection effect is that no matter how hard it is for human-level intelligence to evolve, 100% of evolved civilizations will find themselves originating from planets where it happened anyway. … every newly evolved civilization will find that evolution managed to produce its ancestors." [39]. So even a successful evolutionary run, with fixed computational resources, does not indicate that used compute would be sufficient in a similar experiment, as subsequent runs may not produce similar results. As Shulman and Bostrom put it, "However, reliable creation of human-level intelligence through evolution might require trials on many planets in parallel, with Earth being one of the lucky few to succeed." [39]. Conceivably, "Evolution requires extraordinary luck to hit upon a design for human-level intelligence, so that only 1 in $10^{1000}$ planets with life does so." [39]. Hanson elaborates, "Many have recognized that the recent appearance of intelligent life on Earth need not suggest a large chance that similarly intelligent life appears after a similar duration on any planet like Earth. Since Earth's one data point has been subject to a selection effect, it is consistent with any expected time for high intelligence to arise beyond about a billion years. Few seem to have recognized, however, that this same selection effect also allows the origin of life to be much harder than life's early appearance on Earth might suggest." [76].

EAT attempts to estimate computational requirements theoretically necessary to run different variants of the Darwinian algorithm. Such estimates are usually made with respect to the size of the input problem, which is difficult to formalize with respect to software generation. "It is difficult to characterize the complexity of a problem specific to a method of programming. Holding all things constant, you measure what must change as the size of the input instance increases. It is even more difficult to describe the complexity of a problem that can be solved by a program that is *itself* the output of a program, as is the case with the typical GP. In general, this type of question cannot be answered. What *can* be done however, is to compare the information content of a program with the information content of its output and in this way provide a bound on the complexity of that output." [36].

Specifically, "Though it is impossible to classify the complexity of a problem that can be solved by the output program in advance, it *is* possible to relate the amount of information contained in the output program to the GP itself. By applying the theorems from Kolmogorov complexity, it can be shown that the complexity of the output program of a GP using a pseudo random number generator (PRNG) can be bound above by the GP itself.

**Theorem 3:** For all strings x,y, if x is the shortest program that outputs y, that is K(y)=|x|, then K(x) ≥ K(y) + c.

**Proof:** Let x be the shortest program (by definition, incompressible) that outputs y. That is, K(y) = |x|. Suppose K(y) > K(x). By substitution, |x| > K(x), which is impossible since x was defined as incompressible." [36].

Next, we attempted to include best estimates for Darwinian complexity found in literature. "The performance of an EA is measured by means of the number of function evaluations *T* it makes until an optimal solution is found for the first time. The reason is that evolutionary algorithms tend to be algorithmically simple and each step can be carried out relatively quick. Thus, a function evaluation is assumed to be the most costly operation in terms of computation time. Most often, results about the expected optimization time E(*T*) as a function of *n* are derived where *n* is a measure for the size of the search space. If a fixed-length binary encoding is used *n* denotes the length of the bit strings (and the size of the search space equals $2^n$)." [68]. "[W]ith random mutations, random point mutations, we will get to fitness BB(N) in time exponential in N (evolution by exhaustive search)" [9]. There Busy Beaver function BB(N) = the largest integer that can be named by an N-bit program.

Fitness function evaluation is the most costly procedure in the Darwinian algorithm and is particularly ill defined in the case of software evaluation. How does one formalize a fitness function for something like an operating system, without having to include human users as evaluators? One may be required to rely on Human-Based Genetic Algorithms (HBGA) [77], which would greatly increase time necessary to evaluate every generation and by extension overall simulation time for the run, making it impossible to recapitulate evolution through EAs.

"Essentially, the complexity of an optimization problem for a GA is bound above by the growth rate of the smallest representation [Minimum Chromosome Length - (MCL)] that can be used to solve the problem … . This is because the probabilistic convergence time will remain fixed as a function of the search space. All things held constant, the convergence time will grow as the search space grows." [36]. "This means that the size of the search space doubles for every increase in instance size because the number of possible solutions is equivalent to the number 2 raised to the length of the chromosome, $2^l$." [36]. "By creating a UGP [Universal Genetic Program], we have a single vehicle capable of evolving any program evolvable by a GP. To do this, we treat the first part of the data for the UGP as the *specification* (i.e. the "target" function) for a unique GP. In this way, we can implement *any GP*. This does not eliminate the Kolmogorov complexity bound, rather it determines the *hidden constant* in the Kolmogorov complexity bound." [36]. "Because the output complexity includes all individuals from all populations, producing more individuals through larger populations or longer runs must eventually stop producing new solutions because these solutions would necessarily increase the output complexity beyond the finite limit imposed by the GP." [36]. Others have attempted to calculate total "Computational requirements for recapitulating evolution through genetic algorithms" [39].

Given estimates of computational power of biological evolution in the wild and theoretical analysis for computational resources necessary to run a Darwinian algorithm, we will now try to see if matching compute is currently available, and if not how soon it is predicted to be developed. Currently, world's top 10 supercomputers[1] range from 10 to 125 peta (10^15) floating point operations per second (FLOPS) of theoretical peak performance. For comparison, Bitcoin network[2] currently performs around 35 exa (10^18) hashes per second, many thousands of times the combined speed of the top 500 supercomputers. Similarly, "Storing the total amount of information encoded in DNA in the biosphere, $5.3 \times 10^{31}$ megabases (Mb), would require approximately $10^{21}$ supercomputers with the average storage capacity of the world's four most powerful supercomputers." [73]. "In recent years it has taken approximately 6.7 years for commodity computers to increase in power by one order of magnitude. Even a century of continued

---

[1] https://en.wikipedia.org/wiki/TOP500#Top_10_ranking
[2] https://blockchain.info

Moore's law would not be enough to close this gap. Running more or specialized hardware, or longer runtimes, could contribute only a few more orders of magnitude." [39].

In this section, we looked at estimated computational power of biological evolution and theoretical computational complexity of Darwinian algorithm. In both cases, we found that required computational resources greatly exceed what is currently available and what is projected to be available in the near future. In fact, depending on some assumptions we make with regard to multiverse [78], quantum aspects of biology [79] and probabilistic nature of Darwinian algorithm such compute may never be available. Mahoney arrives at a similar realization: "The biosphere has on the order of $10^{31}$ cells (mostly bacteria) … with $10^6$ DNA base pairs each, encoding $10^{37}$ bits of memory. Cells replicate on the order of $10^6$ seconds, for a total of $10^{48}$ copy operations over the last 3 billion years. If we include RNA transcription and protein synthesis as computing operations, then the evolution of humans required closer to $10^{50}$ operations. By contrast, global computing power is closer to $10^{20}$ operations per second and $10^{22}$ bits of storage. If we were to naively assume that Moore's Law were to continue increasing computing power by a factor of 10 every 5 years, then we would have until about 2080 before we have something this powerful." [80]. Others agree, "The computing resources to match historical numbers of neurons in straightforward simulation of biological evolution on Earth are severely out of reach, even if Moore's law continues for a century. The argument from evolutionary algorithms depends crucially on the magnitude of efficiency gains from clever search, with perhaps as many as thirty orders of magnitude required." [39]. If "… one would have to simulate evolution on vast numbers of planets to reliably produce intelligence through evolutionary methods, then computational requirements could turn out to be many, many orders of magnitude higher still …" [39]. It is hoped by some that future developments in Quantum Evolutionary Computation [81] will help to overcome some of the resource limitations [82] without introducing negative side-effects [83].

## 6. Conclusions

Our analysis of relevant literature shows that no one has succeeded at evolving non-trivial software from scratch, in other words the Darwinian algorithm works in theory, but does not work in practice, when applied in the domain of software production. The reason we do not evolve software is that the space of working programs is very large and discreet. While hill-climbing-heuristic-based evolutionary computations are excellent at solving many optimization problems they fail in the domains of non-continues fitness [84]. This is also the reason we do not evolve complex alife or novel engineering designs. With respect to our two predictions, we can conclude that 1) Simulations of evolution do not produce comparably complex artifacts. 2) Running evolutionary algorithms longer leads to progressively diminishing results. With respect to the three falsifiability conditions, we observe that all three are true as of this writing. Likewise, neither the longest running evolutionary algorithm nor the most complex evolved algorithm nor the most complex digital organism are a part of our common cultural knowledge. This is not an unrealistic expectation as successful software programs, like Deep Blue [85] or Alpha Go [86, 87], are well known to the public.

Others have come to similar conclusions: "It seems reasonable to assume that the number of programs possible in a given language is so inconceivably large that genetic improvement could surely not hope to find solutions in the 'genetic material' of the existing program. The test input space is also, in the words of Dijkstra, "so fantastically high" that surely sampling inputs could never be sufficient to capture static truths about computation." [17]. "... computing science is — and will always be — concerned with the interplay between mechanized and human symbol manipulation, usually referred to as 'computing' and 'programming' respectively. An immediate benefit of this insight is that it reveals "automatic programming" as a contradiction in terms." [88]. Moreover, more specifically, "Genetic algorithms do not scale well with complexity. That is, where the number of elements which are exposed to mutation is large there is often an exponential increase in search space size. This makes it extremely difficult to use the technique on problems such as designing an engine, a house or plane. In order to make such problems tractable to evolutionary search, they must be broken down into the simplest representation possible. Hence we typically see evolutionary algorithms encoding designs for fan blades instead of engines, building shapes instead of detailed construction plans, and airfoils instead of whole aircraft designs. The second problem of complexity is the issue of how to protect parts that have evolved to represent good solutions from further destructive mutation, particularly when their fitness assessment requires them to combine well with other parts." [89].

Even Koza himself acknowledges that it would be highly surprising if his approach could work: "Anyone who has ever written and debugged a computer program and has experienced their brittle, highly non-linear, and perversely unforgiving nature will probably be understandably skeptical about the proposition that the biologically motivated process sketched above could possibly produce a useful computer program." [16]. We challenge EA community to prove us wrong by producing an experiment, which evolves non-trivial software from scratch and without human help. That would be the only way in which our findings could be shown to be incorrect.

On a positive side, the fact that it seems impossible to evolve complex software implies that we are unlikely to be able to evolve highly sophisticated artificially intelligent agents, which may present significant risk to our safety and security [90-96]. Just imagine what would have happened, if the very first time we ran a simulation of evolution on a computer, it produced a superintelligent agent. Yampolskiy has shown that Programming as a problem is AI Complete [97], if GP can solve Programming that would imply that GP = AGI (Artificial General Intelligence), but we see no experimental evidence for such claim. In fact, it is more likely that once we have AGI, it could be used to create an intelligent fitness function for GP and so evolve software. Genetic Programming will not be the cause of Artificial Intelligence, but a product of it.

## Acknowledgments

The author is grateful to Elon Musk and the Future of Life Institute and to Jaan Tallinn and Effective Altruism Ventures for partially funding his work. This is an early draft of this work and I am sure this paper will evolve.